\documentclass[australian,10pt,twocolumn,letterpaper,twocolumn]{article}
\usepackage{lmodern}

\usepackage[T1]{fontenc}
\usepackage[latin9]{inputenc}
\usepackage{xcolor}
\usepackage{bm}
\usepackage{amsmath}
\usepackage{amssymb}
\usepackage{graphicx}
\usepackage[numbers,numbers,sort,compress]{natbib}

\makeatletter

\providecommand{\tabularnewline}{\\}

\usepackage{wacv}
\usepackage{times}
\usepackage{amsmath}
\usepackage{amssymb}
\wacvfinalcopy 
\def\wacvPaperID{13} 
\ifwacvfinal\fi

\usepackage{enumitem}
\setlist[description]{itemsep=0pt,leftmargin=0pt}

\newcommand{\ourmodel}{MargiPose}

\makeatletter
\renewcommand{\paragraph}{%
  \@startsection{paragraph}{4}%
  {\z@}{0.7ex \@plus 0.2ex \@minus 0.2ex}{-1em}%
  {\normalfont\normalsize\bfseries}%
}

\@ifundefined{showcaptionsetup}{}{%
 \PassOptionsToPackage{caption=false}{subfig}}
\usepackage{subfig}
\makeatother

\usepackage{babel}
\begin{document}
\title{3D Human Pose Estimation with 2D Marginal Heatmaps}
\author{Aiden Nibali\hspace{1cm}Zhen He\hspace{1cm}Stuart Morgan\hspace{1cm}Luke
Prendergast\\
La Trobe University, Australia\\
\texttt{\small{}anibali@students.latrobe.edu.au}}

\maketitle
\ifwacvfinal\thispagestyle{empty}\fi
\begin{abstract}
Automatically determining three-dimensional human pose from monocular
RGB image data is a challenging problem. The two-dimensional nature
of the input results in intrinsic ambiguities which make inferring
depth particularly difficult. Recently, researchers have demonstrated
that the flexible statistical modelling capabilities of deep neural
networks are sufficient to make such inferences with reasonable accuracy.
However, many of these models use coordinate output techniques which
are memory-intensive, not differentiable, and/or do not spatially
generalise well. We propose improvements to 3D coordinate prediction
which avoid the aforementioned undesirable traits by predicting 2D
marginal heatmaps under an augmented soft-argmax scheme. Our resulting
model,\emph{ \ourmodel{}}, produces visually coherent heatmaps whilst
maintaining differentiability. We are also able to achieve state-of-the-art
accuracy on publicly available 3D human pose estimation data.
\end{abstract}

\section{Introduction}

Capturing the three-dimensional locations of a person's skeletal joints
has varied applications in areas such as animation~\citep{pullen2002motion},
video game control~\citep{rhodin2014interactive}, and physical rehabilitation~\citep{chang2011physrehab}.
Current standard practice is to use specialised equipment to acquire
the pose of human subjects, such as wearable motion sensors, depth
cameras (\eg Microsoft Kinect~\citep{shotton2011kinect}), or marker-based
motion capture systems. Specialised equipment can be expensive and
restrictive, requiring a specific set up and calibration to be effective.
Recently, effort has gone into constructing models which infer joint
locations using only monocular RGB images taken with a standard camera.
Such models can dramatically improve the accessibility and usability
of pose estimation technology.

Inferring the three-dimensional pose of a human subject from a monocular
image is an inherently under-constrained problem, with the primary
source of ambiguity being a lack of explicit depth information in
the image. However, humans are able to manually recreate the three-dimensional
pose depicted in a photograph by drawing upon a wealth of prior knowledge
and ``intuition'' about visual depth cues, permissible joint rotations,
and likely limb lengths~\citep{marinoiu2016pictorial}. Engineering
an algorithm by hand to emulate this behaviour explicitly has proven
much more difficult.

\begin{figure}
\begin{centering}
\includegraphics[width=0.9\linewidth]{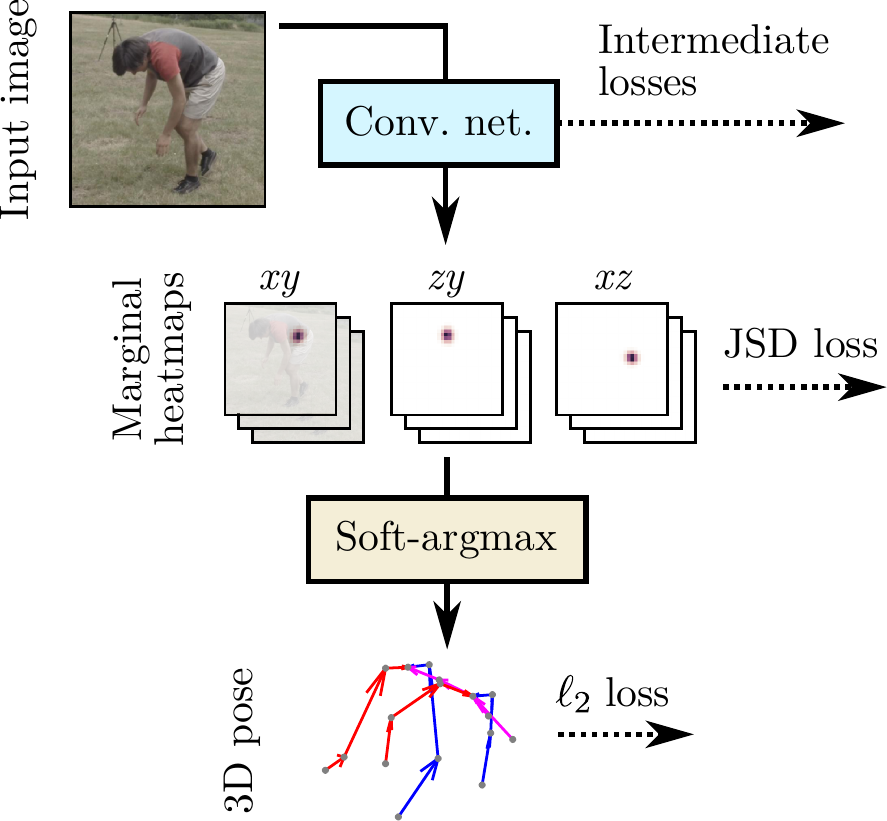}
\par\end{centering}
\caption{\label{fig:overview}High-level system overview for \ourmodel{}.}
\end{figure}
In contrast to hand-crafted algorithms, deep learning approaches are
able to implicitly learn rich statistical relationships between input
and output data. This makes it possible to address the under-constrained
nature of 3D pose estimation by exploiting patterns contained within
training examples, partially encoding the elusive ``prior knowledge''
possessed by humans about likely poses into model weights. Systems
based on deep convolutional neural networks (CNNs) currently achieve
state-of-the-art performance for both 2D and 3D pose estimation, as
is evidenced by results for popular benchmark datasets such as MPII
Human Pose~\citep{andriluka20142d} and Human3.6M~\citep{ionescu2014human3}.

Most existing CNN-based models for 3D pose estimation use either fully
connected output layers or volumetric heatmaps to form joint location
predictions. Fully connected layers lack the inherent spatial equivariance
required for good generalisation, and volumetric heatmaps can be memory-intensive
to produce.

We propose \emph{\ourmodel{}} (Figure~\ref{fig:overview}), a highly
effective CNN-based approach to 3D pose estimation. In contrast to
existing solutions, our model uses 2D marginal heatmaps (in $xy$,
$xz$, and $zy$ space) to predict joint locations. The term ``marginal
heatmap'' is an allusion to the marginalisation of the trivariate
probability mass function represented by a volumetric heatmap.

\ourmodel{}'s architecture is specifically designed to produce marginal
heatmaps from monocular 2D RGB input, accounting for changes in coordinate
space where appropriate. Since it does not produce memory-intensive
volumetric heatmaps, the memory requirement of our model is comparable
to that of 2D pose estimation models. Numerical coordinate values
are calculated from the heatmaps using soft-argmax~\citep{chapelle2010gradient,yi2016lift,luvizon2017human}
with a regularisation term which improves the coherence and interpretability
of learnt heatmap representations. Through experimental evaluation
we find that \ourmodel{} achieves state-of-the-art 3D pose estimation
results on the MPI-INF-3DHP dataset and is highly competitive on the
Human3.6M dataset.

\smallskip{}

\paragraph{Our contributions.}

Firstly, we extend the heatmap-based output strategies commonly found
in 2D pose estimation to the task of 3D pose estimation by predicting
three two-dimensional marginal heatmaps per joint. This is more memory
efficient than the existing heatmap-based 3D pose estimation approaches
which represent heatmaps with volumetric activations.

Secondly, we show that adding a regularisation term to minimise divergence
from ideal Gaussian heatmaps leads to improved accuracy when using
soft-argmax 3D joint location prediction. Furthermore, the resultant
heatmaps are much more visually coherent and readily interpretable
than those predicted by a model trained without regularisation.

Thirdly, we implement a novel CNN model architecture for monocular
3D human pose estimation\footnote{PyTorch code available at https://github.com/anibali/margipose}
and perform evaluation to demonstrate state-of-the-art results on
public datasets. We introduce a technique called \emph{axis permutation}
which manipulates activations in order to account for the discrepancy
between input and output spaces.

\section{Related work}

\subsection{2D human pose estimation}

DeepPose~\citep{toshev2014deeppose} is one of the earliest CNN-based
models for human pose estimation. The model architecture is mostly
convolutional, with fully-connected output layers used to predict
the joint coordinates directly as numerical values. DeepPose features
a cascaded design, where predicted coordinates are repeatedly refined
to produce more accurate results. The basic concept of cascading has
carried forward into more recent 2D pose estimation architectures,
such as Stacked Hourglass~\citep{newell2016stacked}. The Stacked
Hourglass architecture makes extensive use of residual connections~\citep{he2016deep},
and is segmented into multiple stages which permit intermediate supervision
to guide training. The high accuracy of Stacked Hourglass\textemdash as
demonstrated by practical results on the MPII Human Pose dataset~\citep{andriluka20142d}\textemdash has
inspired subsequent work to build upon the architecture directly~\citep{yang2017learning,chen2017advposenet,chou2017self,chu2017multi}.

The overwhelming majority of 2D pose estimation models developed after
DeepPose (including Stacked Hourglass) have ceased using fully-connected
output, instead favouring the more accurate heatmap matching approach~\citep{tompson2014joint}.
Heatmap matching works by training a model to output spatial maps
which indicate joint locations with high-valued pixels. During training,
loss is calculated using the mean squared error between the output
heatmap and an ideal target, 2D spherical Guassian mean-centred on
the ground truth joint location. During inference, numerical coordinates
are calculated from heatmaps through the use of a non-differentiable
argmax operation. In some implementations, the pixels neighbouring
the highest-valued pixel are also considered in order to make minor
sub-pixel adjustments for higher precision results~\citep{newell2016stacked}.

Recently, soft-argmax~\citep{chapelle2010gradient,yi2016lift} has
emerged as a differentiable alternative to argmax for heatmap-based
human pose estimation models~\citep{luvizon2017human}. Since it
is possible to backpropagate through the soft-argmax operation, a
distance-based loss is applied directly to the predicted coordinates.

\subsection{Monocular 3D human pose estimation}

In 2D human pose estimation, coordinates predicted by the model are
in the same $xy$ coordinate space as the input, making it straightforward
to construct a simple fully convolutional network which maps RGB inputs
to $xy$ heatmaps. However, for 3D pose estimation the output coordinates
exist in a space with one more dimension than the input ($xyz$-space);
an $xy$-space heatmap does not encode enough information to recover
the depth of a pose joint.

Broadly speaking, researchers have reacted to this by either resorting
to fully-connected output layers, or devising new ways of generating
and utilising heatmaps in the context of 3D coordinate prediction.

\paragraph{Fully-connected output.}

\noindent There are many 3D pose estimation systems which use fully-connected
output layers~\citep{lin2017recurrent,martinez2017simple,tekin2017learning}.
Although conceptually simple, the dense connections of fully connected
layers undermine spatial equivariance in convolutional neural networks,
which can hinder generalisation. This is exemplified by the inferior
performance of such approaches in the context of 2D pose estimation.
It is therefore desirable to explore techniques beyond fully-connected
output for 3D pose estimation.

\paragraph{Volumetric heatmaps.}

\noindent One way of extending the heatmap notion to 3D is by designing
a model which produces volumetric heatmaps. A disadvantage of this
approach is that the addition of another dimension to the spatial
activations increases the memory requirements of the system considerably.
\citet{pavlakos2017coarse} partially mitigate the impact on memory
consumption by gradually building up the depth resolution of the activations
throughout the network in a coarse-to-fine fashion. \citet{luvizon20182d}
use the soft-argmax operation to calculate coordinates from volumetric
heatmaps. In contrast to argmax, soft-argmax permits the use of low
resolution heatmaps, which in turn lowers overall memory consumption
for the model.

\paragraph{Location maps.}

\noindent \citet{mehta2017vnect} introduce the concept of ``location
maps'', which are spatial representations of location where each
pixel contains an estimate of a particular coordinate's value. For
the $z$-coordinate, this is conceptually similar to a depth map.
Since the location maps are 2D, they can be produced by models that
are less memory-intensive than for volumetric heatmaps. Unfortunately,
evaluation on the Human3.6M dataset revealed that such an approach
is not competitive with the accuracy of existing techniques, including
volumetric heatmaps.

\smallskip{}

We therefore propose marginal heatmaps as an alternative output strategy
for 3D coordinate prediction. Marginal heatmaps are two-dimensional,
and hence do not require memory-intensive volumetric activations.
Furthermore, we find that marginal heatmaps can be used to generate
predictions on benchmark datasets that are highly competitive with
the state-of-the-art.

\section{\label{sec:heatmaps}Marginal heatmaps with soft-argmax}

In this section we will derive marginal heatmaps for use in 3D pose
estimation from the volumetric heatmaps used by existing work~\citep{luvizon20182d,pavlakos2017coarse}.

\paragraph{Volumetric heatmaps.}

Let $X$, $Y$, and $Z$ be random variables corresponding to the
$x$-, $y$-, and $z$-coordinates of the predicted location for a
particular human pose joint in 3D space. Under the soft-argmax prediction
strategy for pose estimation~\citep{luvizon2017human}, the estimated
location of the joint, $\bm{\mu}$, is taken to be the expectation
of the random variables. That is, $\mu_{x}=\mathbb{E}\left[X\right]$,
$\mu_{y}=\mathbb{E}\left[Y\right]$, and $\mu_{z}=\mathbb{E}\left[Z\right]$. 

If we constrain $X$, $Y$, and $Z$ to only cover the values of discrete
locations within a cuboid, we can represent the trivariate probability
mass function $P(X=x,Y=y,Z=z)$ with a volumetric heatmap. The volumetric
heatmap, $\hat{\bm{\mathsf{H}}}$, is a $depth\times height\times width$
tensor where each element represents the probability that the joint
is located at the corresponding spatial location. All elements of
the heatmap must be non-negative and sum to one in order to define
a valid probability mass function.

Equation~\ref{eq:3d_soft_argmax} describes how the estimated location
may be computed from $\hat{\bm{\mathsf{H}}}$ under the soft-argmax
scheme. The coordinate indicator tensors $\bm{\mathsf{\mathcal{X}}}$,
$\bm{\mathsf{\mathcal{Y}}}$, and $\bm{\mathsf{\mathcal{Z}}}$ consist
of elements referencing their own $x$, $y$, and $z$ coordinates
respectively (that is, $\mathcal{X}_{i,j,k}=k$, $\mathcal{Y}_{i,j,k}=j$,
and $\mathcal{Z}_{i,j,k}=i$). We denote the scalar product of vectorised
tensors by $\left\langle \cdot,\cdot\right\rangle $.
\begin{equation}
\mu_{x}=\left\langle \hat{\bm{\mathsf{H}}},\bm{\mathsf{\mathcal{X}}}\right\rangle ,\,\mu_{y}=\left\langle \hat{\bm{\mathsf{H}}},\bm{\mathsf{\mathcal{Y}}}\right\rangle ,\,\mu_{z}=\left\langle \hat{\bm{\mathsf{H}}},\bm{\mathsf{\mathcal{Z}}}\right\rangle \label{eq:3d_soft_argmax}
\end{equation}

\begin{figure}
\begin{centering}
\includegraphics[width=5.5cm]{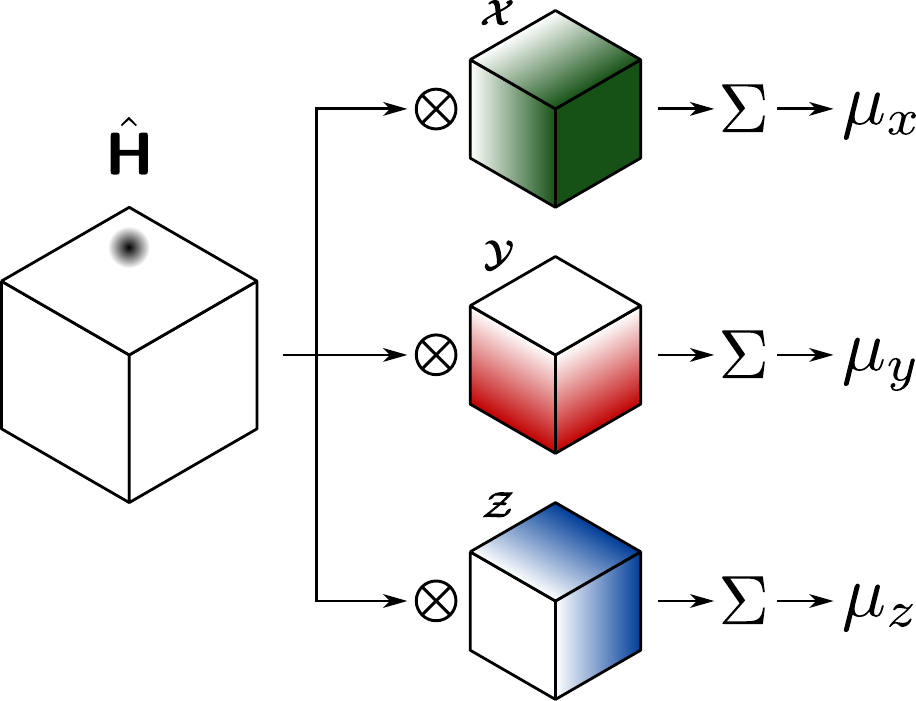}
\par\end{centering}
\caption{\label{fig:3d_soft_argmax}Graphical description of the soft-argmax
calculation for a volumetric heatmap. Darker colours indicate higher
tensor element values.}
\end{figure}
Figure~\ref{fig:3d_soft_argmax} illustrates the 3D soft-argmax calculation
described by Equation~\ref{eq:3d_soft_argmax}. It is a logical extension
of the graphical representation for 2D soft-argmax first presented
in~\citep{luvizon20182d}.

\begin{figure}
\begin{centering}
\includegraphics[width=5cm]{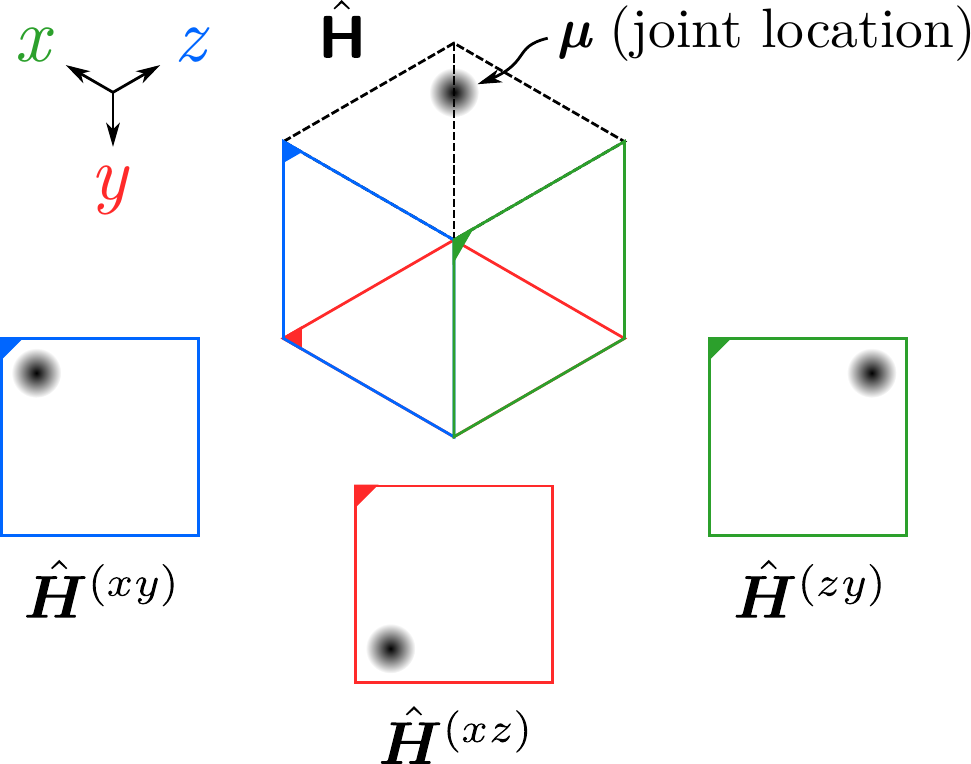}
\par\end{centering}
\caption{\label{fig:hm_views}Multiple two-dimensional ``views'' of a spherical
Gaussian from outside the volumetric heatmap can fully describe the
location of a joint.}
\end{figure}

\paragraph{Marginal heatmaps.}

Volumetric heatmap-based approaches are functional, but three-dimensional
model outputs typically imply high memory usage for neural networks.
This drawback is identified and partially mitigated within existing
work by reducing the resolution of the volumetric heatmaps in some
way~\citep{pavlakos2017coarse,luvizon20182d}. However, it is possible
to circumvent the issue entirely by using marginal heatmaps to avoid
the explicit representation of volumetric heatmaps altogether.

Consider the bivariate marginal probability mass functions $P(X=x,Y=y)$,
$P(Y=y,Z=z)$, and $P(X=x,Z=z)$, and their corresponding two-dimensional
heatmap representations $\hat{\bm{H}}^{(xy)}$, $\hat{\bm{H}}^{(zy)}$,
and $\hat{\bm{H}}^{(xz)}$. Like $\hat{\bm{\mathsf{H}}}$, these marginal
heatmaps are sufficient for calculating the expected values of joint
coordinates. Conceptually, the volumetric and marginal heatmaps are
related as follows:
\begin{equation}
\begin{array}{l}
\hat{\bm{H}}^{(xy)}=\sum_{i}\hat{\bm{H}}_{i,:,:}\\
\hat{\bm{H}}^{(zy)}=(\sum_{k}\hat{\bm{H}}_{:,:,k})^{T}\\
\hat{\bm{H}}^{(xz)}=\sum_{j}\hat{\bm{H}}_{:,j,:}
\end{array}\label{eq:2d_hms}
\end{equation}

More intuitively, the marginal heatmaps may be considered as ``views''
of $\hat{\bm{\mathsf{H}}}$, as illustrated in Figure~\ref{fig:hm_views}.

\begin{table}
\begin{centering}
\setlength\tabcolsep{4pt}
\renewcommand{\arraystretch}{2}%
\begin{tabular}{ccc}
$\mathbb{E}\left[X\right]$ & $\mathbb{E}\left[Y\right]$ & $\mathbb{E}\left[Z\right]$\tabularnewline
\hline 
$\left\langle \hat{\bm{H}}^{(xy)},\bm{\mathcal{X}}_{1,:,:}\right\rangle $ & $\left\langle \hat{\bm{H}}^{(xy)},\bm{\mathcal{Y}}_{:,:,1}\right\rangle $ & $\left\langle \hat{\bm{H}}^{(zy)}{}^{T},\bm{\mathcal{Z}}_{:,:,1}\right\rangle $\tabularnewline
$\left\langle \hat{\bm{H}}^{(xz)},\bm{\mathcal{X}}_{:,1,:}\right\rangle $ & $\left\langle \hat{\bm{H}}^{(zy)}{}^{T},\bm{\mathcal{Y}}_{1,:,:}\right\rangle $ & $\left\langle \hat{\bm{H}}^{(xz)},\bm{\mathcal{Z}}_{:,1,:}\right\rangle $\tabularnewline
\end{tabular}
\par\end{centering}
\caption{\label{tbl:expectation_marginal_heatmaps}The expected value of each
coordinate can be calculated from either of two marginal heatmaps.}
\end{table}
We can use $\mu_{x}=\left\langle \hat{\bm{H}}^{(xy)},\bm{\mathcal{X}}_{1,:,:}\right\rangle $,
or equivalently $\mu_{x}=\left\langle \hat{\bm{H}}^{(xz)},\bm{\mathcal{X}}_{:,1,:}\right\rangle $,
to calculate a soft-argmax estimate for the $x$-coordinate of the
joint location. The $y$- and $z$-coordinates may be estimated similarly
with appropriate heatmaps, as shown in Table~\ref{tbl:expectation_marginal_heatmaps}.
Using this formulation we only require a model which predicts three
2D heatmaps per joint, obviating the need to predict $\hat{\bm{\mathsf{H}}}$
directly. Such a model will typically require less memory than a volumetric
equivalent. This is a notable difference from existing work using
marginal distributions, which still produce three-dimensional heatmaps
as an intermediate step~\citep{luvizon20182d,pavlakos2018ordinal}.

Since the model is free to predict the marginal heatmaps independently,
they will generally not be consistent with one another. That is to
say, there is no guarantee that the rows in Table~\ref{tbl:expectation_marginal_heatmaps}
will resolve to the same values. There are therefore two ways to calculate
each of $\mathbb{E}\left[X\right]$, $\mathbb{E}\left[Y\right]$,
and $\mathbb{E}\left[Z\right]$, depending on which marginal heatmap
is used. We address this by making coordinate predictions according
to Equation~\ref{eq:joint_location_est}. Due to the $xy$ heatmap
having the same orientation as the input image, the $x$- and $y$-coordinates
are predicted using $\hat{\bm{H}}^{(xy)}$ alone. However, the $z$-coordinate
is taken to be the average expectation obtained from the other two
heatmaps to better estimate depth.
\begin{equation}
\begin{array}{c}
\mu_{x}=\left\langle \hat{\bm{H}}^{(xy)},\bm{\mathcal{X}}_{1,:,:}\right\rangle \\
\mu_{y}=\left\langle \hat{\bm{H}}^{(xy)},\bm{\mathcal{Y}}_{1,:,:}\right\rangle \\
\mu_{z}=\frac{1}{2}\left\langle \hat{\bm{H}}^{(zy)}{}^{T},\bm{\mathcal{Z}}_{:,:,1}\right\rangle +\frac{1}{2}\left\langle \hat{\bm{H}}^{(xz)},\bm{\mathcal{Z}}_{:,1,:}\right\rangle 
\end{array}\label{eq:joint_location_est}
\end{equation}

\section{Model architecture}

\begin{figure*}
\begin{centering}
\includegraphics[width=1\linewidth]{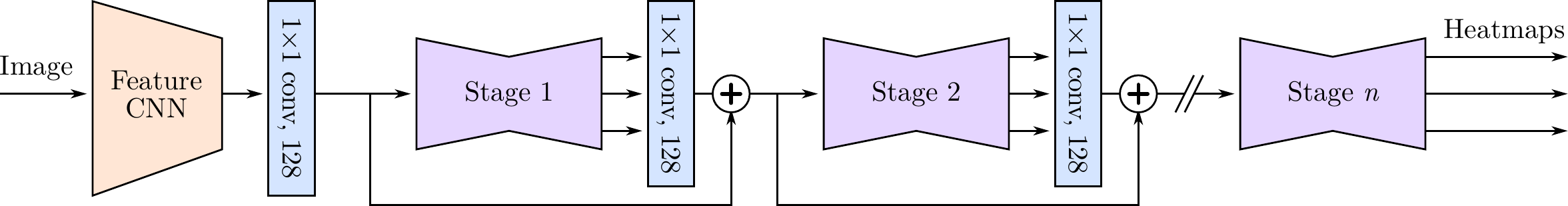}
\par\end{centering}
\caption{\label{fig:complete-model}The complete high-level model architecture.
The internal structure of each stage is detailed in Figure~\ref{fig:chatterbox_stage}.
``Feature CNN'' is a truncated Inception v4 model~\citep{szegedy2017inception}.
Loss is computed at each stage output.}
\end{figure*}
It is necessary to have a model capable of predicting marginal heatmaps
in order to use the prediction strategy outlined in Section~\ref{sec:heatmaps}.
Since pose estimation data is inherently spatial, convolutional layers
are a natural foundation for the model.

The calculation performed by each convolutional layer is spatially
local. That is, for any given output pixel, the value of that pixel
is calculated using input pixels that are within a fixed spatial neighbourhood.
This is appropriate when both the input and output images exist in
the same coordinate space and there is a correlation between the locations
of input and output features. For example, in 2D pose estimation the
output heatmaps and input RGB image both exist in $xy$ coordinate
space, and the ground truth target spherical Gaussians align with
the joints in the input image. 

However, we require our model to not only output an $xy$ heatmap,
$\hat{\bm{H}}^{(xy)}$, but also heatmaps that have one axis in the
$z$-direction, $\hat{\bm{H}}^{(zy)}$ and $\hat{\bm{H}}^{(xz)}$.
This poses a challenge for convolution-based computation. Consider
the case of predicting a heatmap in the $zy$-plane, $\hat{\bm{H}}^{(zy)}$,
from an input image in the $xy$-plane. In general, a location in
the $z$-direction does not correspond to a location in the $x$-direction.
This means that there may be quite some distance between visual evidence
in the input image and the desired prediction location in the output
image. Such an arrangement is generally not ideal for convolutional
neural networks.

For 3D pose estimation, the spatial discrepancy is never along both
axes at once (Table~\ref{tbl:axes_correspondences} shows axis correspondences
for each of the three heatmaps). It is therefore desirable to preserve
spatial locality of computation along the appropriate axes.

\begin{table}
\begin{centering}
\begin{tabular}{|c|c|c|}
\hline 
 & \multicolumn{2}{c|}{Correspondence with input}\tabularnewline
\hline 
Heatmap & Horizontal & Vertical\tabularnewline
\hline 
\hline 
$\hat{\bm{H}}^{(xy)}$ & \textcolor{teal}{Yes} & \textcolor{teal}{Yes}\tabularnewline
\hline 
$\hat{\bm{H}}^{(xz)}$ & \textcolor{teal}{Yes} & \textcolor{red}{No}\tabularnewline
\hline 
$\hat{\bm{H}}^{(zy)}$ & \textcolor{red}{No} & \textcolor{teal}{Yes}\tabularnewline
\hline 
\end{tabular}
\par\end{centering}
\caption{\label{tbl:axes_correspondences}Image axes correspondences between
different output heatmap types and the input image in $xy$-space.}
\end{table}
\begin{figure}
\begin{centering}
\includegraphics[width=1\linewidth]{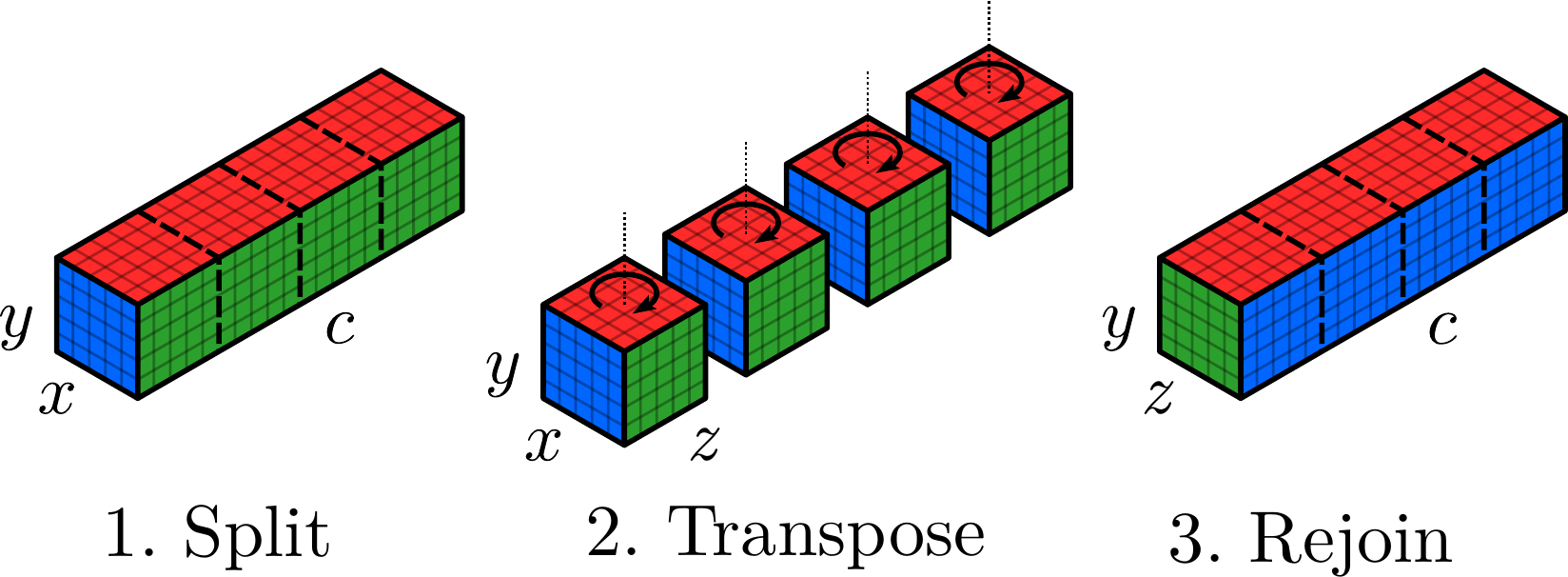}
\par\end{centering}
\caption{\label{fig:permute}Using axis permutation on activations to transition
from $xy$ to $zy$ space.}
\end{figure}

\paragraph{Axis permutation.}

\noindent By transposing the intermediate activations it is possible
to permute the axis undergoing spatially-local calculations with the
axis undergoing densely connected calculations. Therefore the model
can be built using convolutional layers without depending on spatial
correspondence between mismatched axes. This allows the model to aggregate
depth cues into feature maps, which will then become pixel values
along the $z$-axis. Figure~\ref{fig:permute} illustrates the axis
permutation operation for $\hat{\bm{H}}^{(zy)}$. Note that the permutation
operation is simply a fixed manipulation of the activations, and does
not add any parameters to the model.

\paragraph{Overall model architecture.}

\begin{figure}
\begin{centering}
\includegraphics[width=0.9\linewidth]{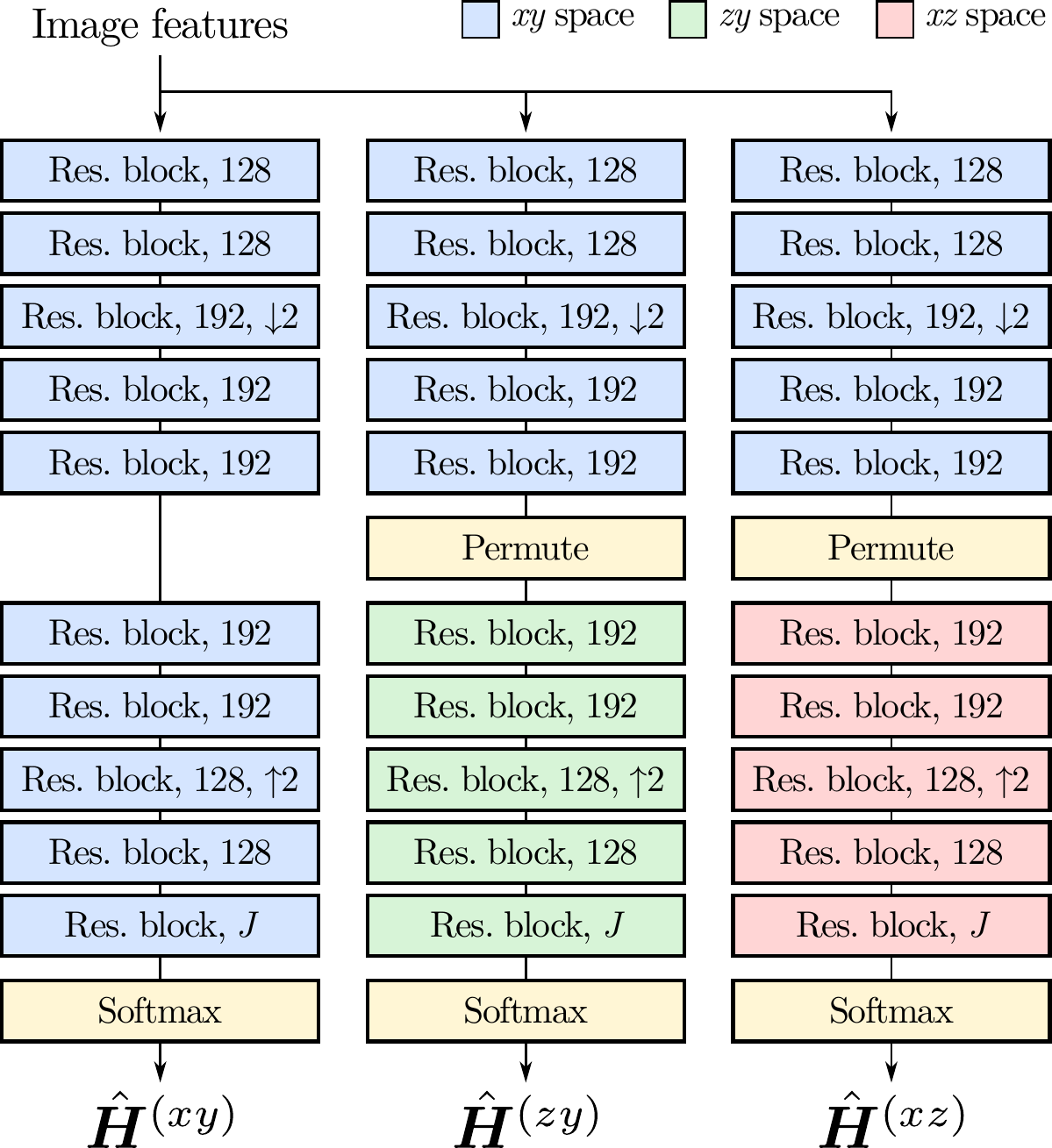}
\par\end{centering}
\caption{\label{fig:chatterbox_stage}The internal structure of a heatmap prediction
stage. Residual blocks are labelled with the number of output channels.}
\end{figure}
Figure~\ref{fig:chatterbox_stage} illustrates the arrangement of
residual blocks we used to produce heatmaps from image features. Residual
blocks are constructed as per ResNet using ``option C'' shortcut
connections~\citep{he2016deep}. For the network paths predicting
$\hat{\bm{H}}^{(zy)}$ and $\hat{\bm{H}}^{(xz)}$, the axis permutation
operation is applied mid-way through the stage.

The complete model is assembled according to Figure~\ref{fig:complete-model}.
Features are extracted from $256\times256$ pixel input images using
a truncated Inception v4 model \citep{szegedy2017inception}. Multiple
heatmap prediction stages are stacked together after the feature extractor
to increase the capacity of the model. ``Adapter'' $1\times1$ convolution
layers are placed in between the stages to combine the previous heatmap
predictions into feature maps, which are added with the previous stage's
input to form a large skip connection. This stacking technique is
inspired by the Stacked Hourglass architecture for 2D pose estimation~\citep{newell2016stacked}.

\section{Joint location loss}

The typical heatmap matching with argmax approach used by most current
2D pose estimation networks has proven to be effective for producing
highly accurate predictions~\citep{tompson2014joint,newell2016stacked}.
The loss function used in this arrangement is simply the mean squared
error between $\hat{\bm{\mathsf{H}}}$ (the predicted heatmap), and
$\bar{\bm{\mathsf{H}}}$, a synthetic heatmap drawn using a Gaussian
centred on the ground truth joint location.
\begin{equation}
\mathcal{L}_{HM}=\left\Vert \hat{\bm{\mathsf{H}}}-\bar{\bm{\mathsf{H}}}\right\Vert _{2}^{2}\label{eq:matching_loss}
\end{equation}

Such a loss function provides a strong training signal to the network
as a result of the pixel-wise output gradients. However, this approach
to coordinate prediction has one major issue\textemdash the actual
location of the joint is estimated using the non-differentiable argmax
operation. This final calculation means that the resulting model is
not entirely differentiable, and hence cannot be built upon and trained
using end-to-end backpropagation. Furthermore, argmax causes the precision
of the coordinates to be limited by the resolution of the heatmap.
This severely impedes models which produce heatmaps at a low spatial
resolution.

Soft-argmax~\citep{luvizon2017human} is a differentiable alternative
to argmax which does not have its precision tied to the resolution
of heatmaps. In contrast to the heatmap matching approach, loss is
applied to the coordinates directly and backpropagated through the
soft-argmax calculation. For example, the $\ell_{2}$ loss between
the predicted joint location ($\bm{\mu}$) and the ground truth joint
location ($\bar{\bm{\mu}}$) may be used, as shown in Equation~\ref{eq:l2_loss}.
\begin{equation}
\mathcal{L}_{\ell_{2}}=\left\Vert \bm{\mu}-\bar{\bm{\mu}}\right\Vert _{2}\,\mathrm{where}\,\bm{\mu}=\mathbb{E}[\hat{\bm{\mathsf{H}}}]\label{eq:l2_loss}
\end{equation}

\begin{figure}
\hspace*{\fill}\subfloat[\label{fig:hm_no_reg}]{\includegraphics[width=0.3\linewidth]{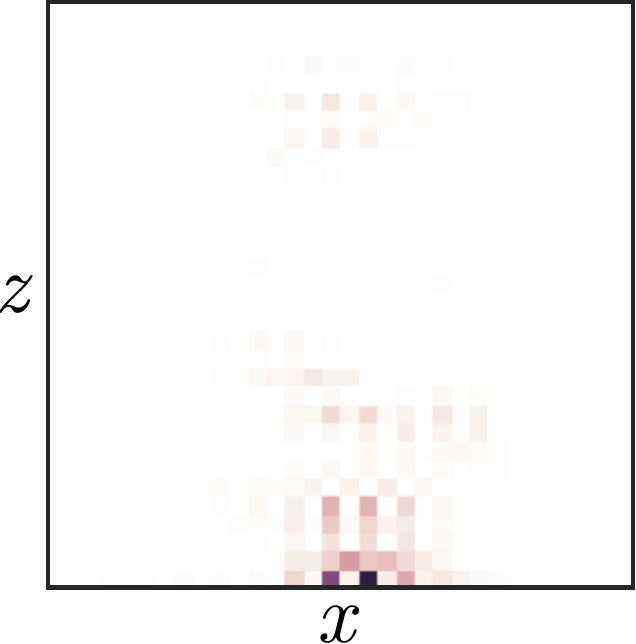}

}\hspace*{\fill}\subfloat[\label{fig:hm_reg}]{\includegraphics[width=0.3\linewidth]{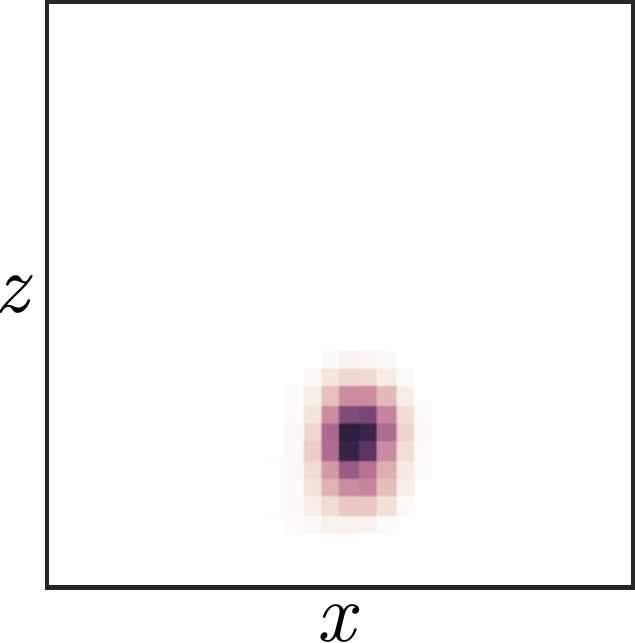}

}\hspace*{\fill}

\caption{Right wrist joint $xz$ heatmaps generated by models trained using
(a) $\ell_{2}$ loss only, and (b) $\ell_{2}$ loss with Jensen-Shannon
divergence regularisation.}
\end{figure}

There are many possible heatmaps which will minimise $\mathcal{L}_{\ell_{2}}$
equally well. Factors such as the ``spread'' or ``shape'' of the
heatmap do not affect the loss, provided that the expected value remains
constant. Initially this might seem beneficial, as the model is free
to represent heatmaps in whichever way facilitates more accurate predictions.
However, in practice the lack of pixel-wise supervision results in
weaker gradients which do not support training particularly well and
can hinder test-set performance. Furthermore the $\ell_{2}$ loss
alone does not result in visually coherent heatmaps, as is shown in
Figure~\ref{fig:hm_no_reg}.

\subsection{Regularised soft-argmax}

\global\long\def\jsd{\operatorname{JSD}}%

We propose to combine the strengths of heatmap matching and soft-argmax
by introducing a regularisation term to the soft-argmax loss function
which explicitly guides the form of predicted heatmaps. Such guidance
dramatically improves the visual coherence of predicted heatmaps,
as is shown in Figure~\ref{fig:hm_reg}. As we will later demonstrate
in our experimental results, introducing such a regularisation term
also significantly improves prediction accuracy.

Let us once again return to the probabilistic interpretation of heatmaps.
If we want to encourage heatmaps to mimic the shape of a specific
probability distribution, we can minimise the the Jensen-Shannon divergence
(JSD)~\citep{lin1991divergence} from that particular distribution.
As long as the mean of the target distribution matches the ground
truth joint location, minimising such a divergence will not compete
with minimising $\mathcal{L}_{\ell_{2}}$. Therefore the two objectives
remain complementary and training is stable.

In accordance with existing work in 2D pose estimation, we use spherical
Gaussians as targets for predicted heatmaps. Equation~\ref{eq:loss_function}
shows the complete hybrid per-joint loss function, where $\bm{\sigma}^{2}$
is the variance of the target Gaussian. We find that setting $\bm{\sigma}=\bm{I}$
(1 pixel) works well for $32\times32$ pixel heatmaps.
\begin{equation}
\begin{array}{rcl}
\mathcal{L}_{3D} & = & \left\Vert \bm{\mu}-\bar{\bm{\mu}}\right\Vert _{2}+\\
 &  & \jsd\left(\hat{\bm{H}}^{(xy)}\parallel\mathcal{N}(\bar{\bm{\mu}}{}_{xy},\bm{\sigma}^{2})\right)+\\
 &  & \jsd\left(\hat{\bm{H}}^{(zy)}\parallel\mathcal{N}(\bar{\bm{\mu}}{}_{zy},\bm{\sigma}^{2})\right)+\\
 &  & \jsd\left(\hat{\bm{H}}^{(xz)}\parallel\mathcal{N}(\bar{\bm{\mu}}{}_{xz},\bm{\sigma}^{2})\right)
\end{array}\label{eq:loss_function}
\end{equation}

For examples which contain only 2D joint annotations, loss is applied
to $\hat{\bm{H}}^{(xy)}$ only. That is,
\begin{equation}
\mathcal{L}_{2D}=\left\Vert \bm{\mu}_{xy}-\bar{\bm{\mu}}_{xy}\right\Vert _{2}+\jsd\left(\hat{\bm{H}}^{(xy)}\parallel\mathcal{N}(\bar{\bm{\mu}}{}_{xy},\bm{\sigma}^{2})\right)\label{eq:2d_loss_function}
\end{equation}

\begin{figure}
\begin{centering}
\includegraphics[height=5.5cm]{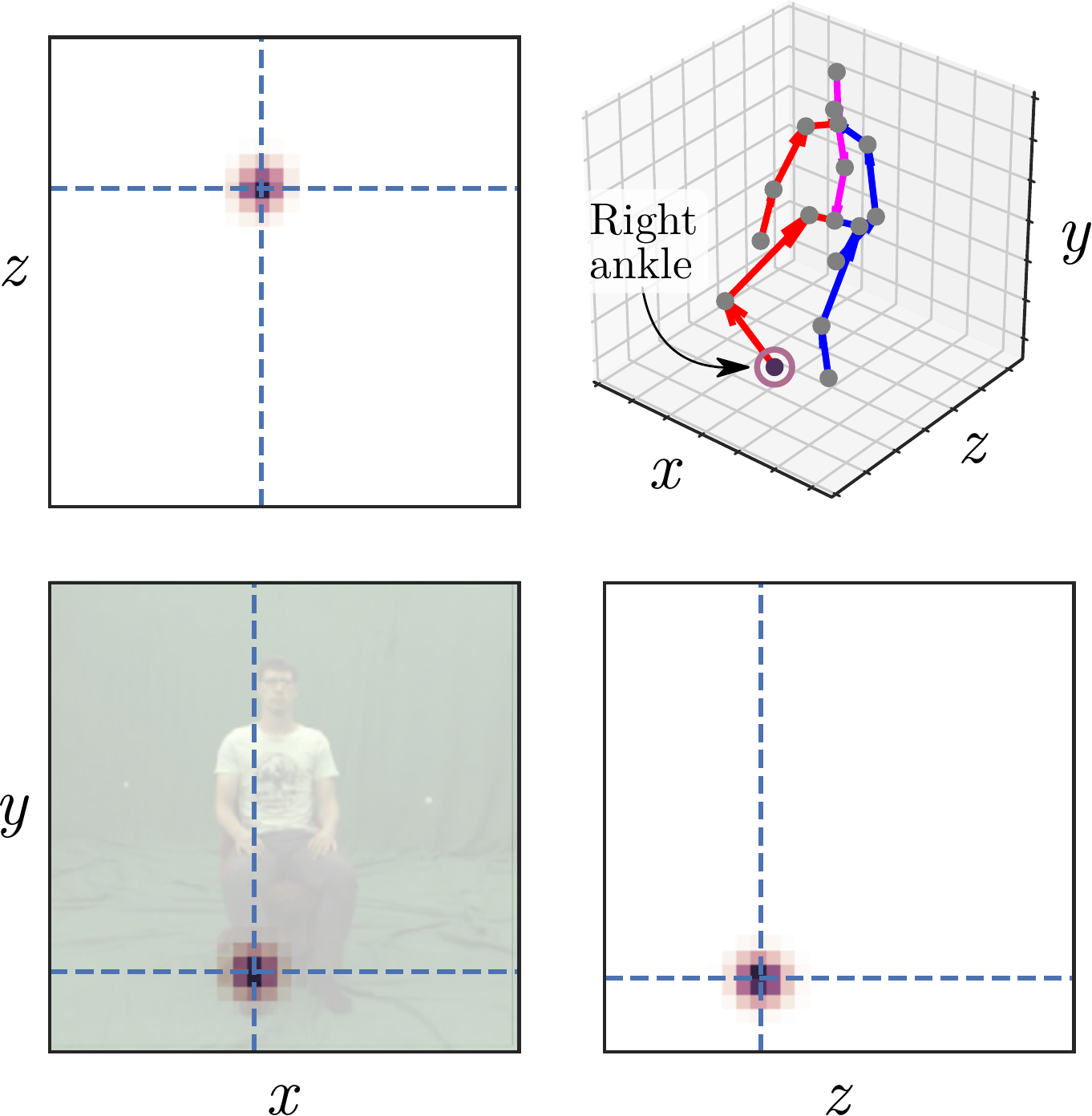}
\par\end{centering}
\caption{\label{fig:rankle_heatmaps}Right ankle heatmaps predicted by \ourmodel{}
for a test set example. The dashed crosshairs indicate the calculated
expectation of the joint location.}
\end{figure}
In practice, the additional divergence-based loss term helps to make
the heatmaps more coherent and interpretable. For example, Figure~\ref{fig:hm_reg}
clearly shows that the heatmap has a greater spread along the $z$-axis,
which expresses that in this particular instance the $z$ coordinate
is less certain than the $x$ coordinate. In contrast, soft-argmax
\emph{without} regularisation (Figure~\ref{fig:hm_no_reg}) makes
it much more difficult to interpret the output of the model.

Figure~\ref{fig:rankle_heatmaps} depicts all three marginal heatmaps
generated for a right ankle joint prediction by \ourmodel{} after
training with JSD regularisation. The effect of the regularisation
is clear\textemdash each heatmap strongly resembles a Gaussian centred
on the joint location.

\paragraph{Applying the loss.}

Since \ourmodel{} has multiple prediction stages, loss is calculated
after each stage. Such intermediate supervision supplies the model
with gradients closer to the input, providing more guidance for training
early layers. The model can be trained on 3D- and 2D-annotated data
simultaneously by switching between $\mathcal{L}_{3D}$ and $\mathcal{L}_{2D}$
on a per-example basis, then aggregating to form a batch loss.

\section{Experiments}

\begin{figure*}
\begin{centering}
\includegraphics[height=3.9cm]{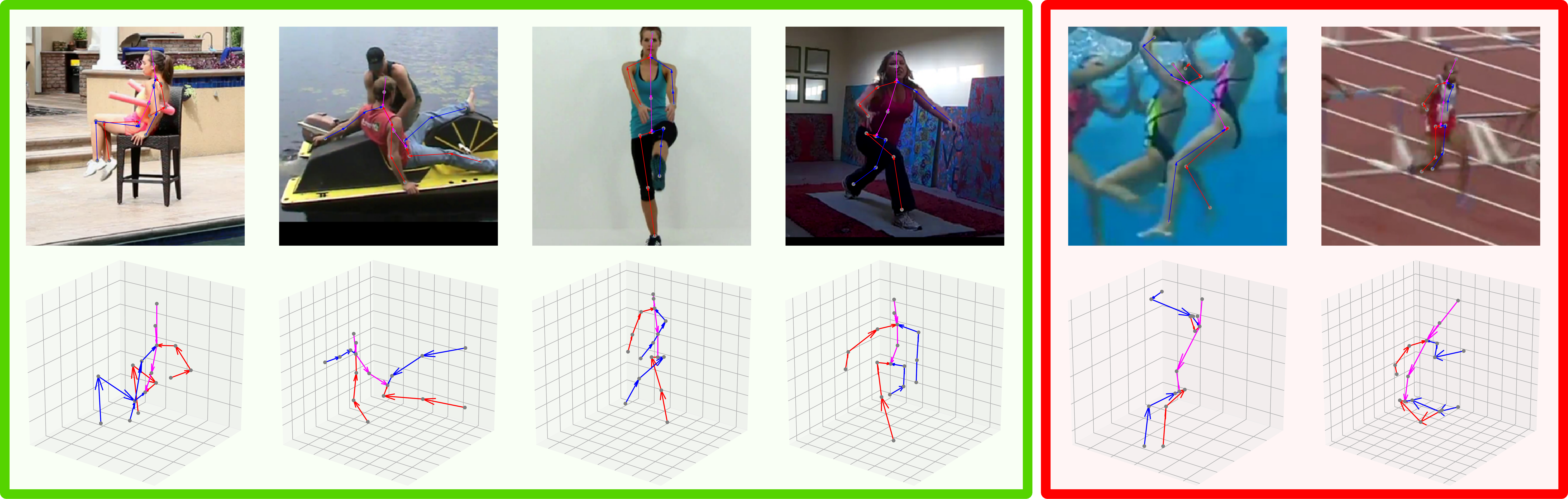}
\par\end{centering}
\caption{\label{fig:mpii_qualitative}Good (left) and poor (right) 3D pose
predictions generated by our model on MPII dataset images.}
\end{figure*}

\subsection{Datasets}

Our experimental evaluation is focussed on two publicly available
datasets for 3D pose estimation: Human3.6M and MPI-INF-3DHP. We also
make use of MPII Human Pose data to augment training.
\begin{description}
\item [{MPII~Human~Pose~\citep{andriluka20142d}}] is a popular 2D pose
estimation dataset comprised of still frames from YouTube videos.
Each image contains at least one human subject that has been manually
annotated with a 16-joint skeleton in 2D.
\item [{Human3.6M~\citep{ionescu2014human3}}] consists of footage recorded
in a lab environment. Each subject's joints were labelled in 3D using
an automated, marker-based system. Human3.6M is very popular as a
3D pose benchmark, but has some important limitations. Firstly, models
trained on this data can exploit the wearable markers as visual cues,
and as such it is difficult to evaluate how well such a model would
generalise to markerless situations. Secondly, Human3.6M is not representative
of real-world scene variety due to each image containing the same
background environment.
\item [{MPI-INF-3DHP~\citep{mehta2017monocular}}] is a recent 3D pose
dataset which overcomes some of the limitations of Human3.6M. All
training video is taken in a lab environment against a green-screen,
but the test set contains a mixture of indoor and outdoor footage.
Hence models must generalise beyond the green-screen lab environment
in order to achieve high accuracy on the test set. Unlike Human3.6M,
subjects in this dataset do not wear visible markers.
\end{description}
Annotations from all three datasets were unified using a canonical
skeleton of 17 joints in order to better facilitate model training.
The joints included in the canonical skeleton are head (top and front),
neck, shoulders, elbows, wrists, hips, knees, ankles, pelvis, and
spine (middle).

\subsection{Data augmentation}

Training examples were dynamically augmented using random adjustments
to scale, position, rotation, colour, and horizontal flipping. Additionally,
compositing was used on images from the MPI-INF-3DHP dataset to vary
the appearance of clothing and backgrounds~\citep{mehta2017monocular,mehta2017vnect}.

For each experimental configuration we report metrics using unaugmented
test set examples. We additionally report results using the ten-crop
test data augmentation scheme of \citet{luvizon20182d}. Such results
are marked explicitly as ``multi-crop''.

\subsection{Training}

Separate models were trained for the Human3.6M and MPI-INF-3DHP datasets
in order to compare against existing work. Each training batch consists
of 32 examples, with 16 samples drawn from the 3D dataset and 16 from
the 2D MPII dataset. The visual diversity in the 2D MPII dataset examples
aids generalisation at inference time.

Stochastic gradient descent with a momentum value of 0.9 was used
to optimise the model parameters. The learning rate was varied according
to the 1-cycle learning rate schedule~\citep{smith2018disciplined},
with $LR_{max}=1.0$.

\subsection{Evaluation protocols}

For the Human3.6M dataset we adopt the popular evaluation protocol
of using subjects S1, S5, S6, S7, and S8 for training, and evaluating
on every $64^{th}$ frame for subjects S9 and S11~\citep{sun2017compositional,luvizon20182d}.
The ground truth root joint depth is used to recover the depth of
the predicted skeleton, which is necessary to break the ambiguity
between depth and scale. The predicted and ground truth skeletons
are then translated so that the root joints align with the origin
before acquiring comparison metrics.

For the MPI-INF-3DHP dataset, we report results using universally-scaled
skeletons (fixed scale of 920 mm knee\textendash neck). Since the
scale is known, the ground truth root joint depth is not used to find
the absolute depth of the predicted skeleton. As with the Human3.6M
dataset, skeletons are translated to the origin before comparison.
We use the same subset of 14 joints as \citet{mehta2017vnect} for
our evaluation (\ie pelvis, spine, and front of head are excluded).
The initial release of the MPI-INF-3DHP dataset used by related works
\citep{mehta2017vnect,dabral2017structure,kanazawa2017end} had systematic
errors in pose labels for test set subjects TS3 and TS4. We evaluate
on the original, erroneous labels in order to compare with results
reported by existing works, but also provide results on the corrected
test set as a reference point for future models.

The main metrics considered are PCK (percentage of correct keypoints),
MPJPE (mean per joint position error), and AUC (area under curve).
PCK measures the percentage of predicted joint locations which are
within 150mm of the ground truth. MPJPE measures the mean $\ell_{2}$
distance between predicted and ground truth joint locations in millimetres.
AUC measures the average PCK over a range of thresholds (0-150mm).

Some of our results are marked as using Procrustes alignment. In these
instances the predicted pose skeletons are compared with the ground
truth up to a similarity transform, which is useful for disentangling
the local configuration of the pose from global positioning within
the scene.

\subsection{Ablative study}

\begin{table}
\begin{centering}
\begin{tabular}{lccc}
\hline 
 & PCK & MPJPE & AUC\tabularnewline
\hline 
Minimal model & 90.1 & 74.9 & 55.0\tabularnewline
Above + axis permutation & 90.4 & 74.5 & 55.3\tabularnewline
Above + regularisation & 92.2 & 68.5 & 58.4\tabularnewline
Above + 4-stage & \textbf{94.3} & \textbf{62.7} & \textbf{61.0}\tabularnewline
\hline 
\end{tabular}
\par\end{centering}
\caption{\label{tbl:ablation}Ablative study results evaluated on the MPI-INF-3DHP
test set with Procrustes alignment.}
\end{table}
\begin{table}
\begin{centering}
\begin{tabular}{ccc}
\hline 
Method & MPJPE & PA MPJPE\tabularnewline
\hline 
\citet{kanazawa2017end} & 88.0 & 58.1\tabularnewline
\citet{rogez2017lcr} & 87.7 & 71.6\tabularnewline
\citet{mehta2017vnect} & 80.5 & -\tabularnewline
\citet{pavlakos2017coarse} & 71.9 & 51.9\tabularnewline
\citet{martinez2017simple} & 62.9 & 47.7\tabularnewline
\citet{sun2017compositional} & 59.1 & 48.3\tabularnewline
\citet{luvizon20182d} (multi-crop) & \textbf{53.2} & -\tabularnewline
\hline 
\ourmodel{} & 57.0 & 40.4\tabularnewline
\ourmodel{} (multi-crop) & 55.4 & \textbf{39.0}\tabularnewline
\hline 
\end{tabular}
\par\end{centering}
\caption{\label{tbl:h36m}Results on the Human3.6M dataset. Metrics are shown
with and without Procrustes alignment (PA).}
\end{table}
An ablative study was conducted to investigate how much each component
of the \ourmodel{} system contributes to pose estimation accuracy.
For these experiments our models were trained for 3.2 million iterations.
Starting with a simple model containing a single heatmap prediction
stage, additional components were enabled in a cumulative fashion.
Evaluation was performed on the MPI-INF-3DHP test set with corrected
labels and Procrustes alignment enabled. Our results are shown in
Table~\ref{tbl:ablation}.

Enabling regularisation for soft-argmax had a very noticeable positive
impact on accuracy, improving PCK by 1.8 percentage points and MPJPE
by 6 mm. This finding provides compelling empirical evidence for the
benefits of combining soft-argmax with a pixel-wise heatmap loss.
Increasing the model capacity by raising the number of heatmap prediction
stages from one to four resulted in an additional performance increase
of similar magnitude.

Axis permutation was shown to be of only minor benefit within the
context of our model. This is likely due to the effective receptive
field of the network being large enough for spatially disparate locations
to be bridged by convolutions.

\subsection{Benchmark dataset results}

We compare the results of our 3D pose estimation model to the results
reported by a range of existing work on the Human3.6M dataset. For
these experiments our models were trained for 4.8 million iterations.
Table~\ref{tbl:h36m} shows that \ourmodel{} achieves the best Procrustes-aligned
MPJPE of all systems which report this metric. This indicates that
accurately inferring the local configuration of pose joints is a strength
of our model. Without Procrustes alignment, \ourmodel{} is still
highly competitive with the state-of-the-art.

\begin{table}
\begin{centering}
\begin{tabular}{cccc}
\hline 
Method & PCK & MPJPE & AUC\tabularnewline
\hline 
\citet{dabral2017structure} & 72.3 & 116.3 & 34.8\tabularnewline
\citet{kanazawa2017end} & 72.9 & 124.2 & 36.5\tabularnewline
\citet{mehta2017vnect} & 75.7 & 117.6 & 39.3\tabularnewline
\hline 
\ourmodel{} & 84.7 & 93.7 & 46.3\tabularnewline
\ourmodel{} (multi-crop) & \textbf{85.4} & \textbf{91.3} & \textbf{47.0}\tabularnewline
\hline 
\end{tabular}
\par\end{centering}
\caption{\label{tbl:mpi3d}Results on the MPI-INF-3DHP dataset (uncorrected
labels) without Procrustes alignment.}

\end{table}
Evaluation results on the MPI-INF-3DHP test set provide a better indication
of model generalisation to real-world scenarios. On the MPI-INF-3DHP
dataset (with the original, uncorrected test set labels), our model
exhibits much higher accuracy than existing approaches (Table~\ref{tbl:mpi3d}).
In particular, \ourmodel{} achieves a full 9.4 percentage points
greater PCK than the next best model.

We also report our results using the updated MPI-INF-3DHP test set
labels in Table~\ref{tbl:mpi3d-corrected}. These results are intended
to provide a baseline for future work to compare against.

\begin{table}
\begin{centering}
\begin{tabular}{cccc}
\hline 
Method & PCK & MPJPE & AUC\tabularnewline
\hline 
\ourmodel{} & 87.6 & 87.6 & 48.8\tabularnewline
\ourmodel{} (multi-crop) & 88.3 & 85.2 & 49.6\tabularnewline
\hline 
\ourmodel{} (PA) & 94.8 & 61.6 & 61.4\tabularnewline
\ourmodel{} (PA, multi-crop) & 95.1 & 60.1 & 62.2\tabularnewline
\hline 
\end{tabular}
\par\end{centering}
\caption{\label{tbl:mpi3d-corrected}Results on the MPI-INF-3DHP dataset (corrected
labels). Metrics are shown with and without multi-crop evaluation
and Procrustes alignment (PA).}
\end{table}

\paragraph{Qualitative results.}

In order to evaluate the ability of \ourmodel{} to generalise to
challenging ``in-the-wild'' images, we generated predictions for
MPII test set examples using the model trained for MPI-INF-3DHP prediction.
Since the MPII dataset does not include 3D annotations, we could not
train a model on MPII data alone. Figure~\ref{fig:mpii_qualitative}
exhibits sample predictions generated by \ourmodel{}. Despite all
of the 3D-annotated training data originating from a laboratory environment,
our model is able to generalise to a wide range of different situations
and poses. Predictions are poorest for images containing high levels
of distortion (\eg motion blur), extreme occlusion, or other people
nearby the subject.

\section{Conclusion}

2D marginal heatmaps are a memory-efficient alternative to volumetric
heatmaps when building models for 3D pose estimation. By extending
the soft-argmax loss function with a regularisation term which guides
the shape of heatmaps, such models can be trained effectively to produce
accurate joint predictions. An interesting direction for future work
would be to make more effective use of examples with 2D-only annotations,
thus improving the ability of \ourmodel{} to learn from varied environments
and poses.

\bibliographystyle{abbrvnat}
\bibliography{shortstrings,references}

\newcommand{\arxiv}[1]{arXiv preprint}
\begin{thebibliography}{33}
\providecommand{\natexlab}[1]{#1}
\providecommand{\url}[1]{\texttt{#1}}
\expandafter\ifx\csname urlstyle\endcsname\relax
  \providecommand{\doi}[1]{doi: #1}\else
  \providecommand{\doi}{doi: \begingroup \urlstyle{rm}\Url}\fi

\bibitem[Andriluka et~al.(2014)Andriluka, Pishchulin, Gehler, and
  Schiele]{andriluka20142d}
M.~Andriluka, L.~Pishchulin, P.~Gehler, and B.~Schiele.
\newblock {2D} human pose estimation: New benchmark and state of the art
  analysis.
\newblock In \emph{Proc. {CVPR}}. IEEE, 2014.

\bibitem[Chang et~al.(2011)Chang, Chen, and Huang]{chang2011physrehab}
Y.-J. Chang, S.-F. Chen, and J.-D. Huang.
\newblock A kinect-based system for physical rehabilitation: A pilot study for
  young adults with motor disabilities.
\newblock \emph{Research in developmental disabilities}, 2011.

\bibitem[Chapelle and Wu(2010)]{chapelle2010gradient}
O.~Chapelle and M.~Wu.
\newblock Gradient descent optimization of smoothed information retrieval
  metrics.
\newblock \emph{Information retrieval}, 2010.

\bibitem[Chen et~al.(2017)Chen, Shen, Wei, Liu, and Yang]{chen2017advposenet}
Y.~Chen, C.~Shen, X.-S. Wei, L.~Liu, and J.~Yang.
\newblock Adversarial {PoseNet}: A structure-aware convolutional network for
  human pose estimation.
\newblock In \emph{Proc. {ICCV}}. IEEE, 2017.

\bibitem[Chou et~al.(2017)Chou, Chien, and Chen]{chou2017self}
C.-J. Chou, J.-T. Chien, and H.-T. Chen.
\newblock Self adversarial training for human pose estimation.
\newblock In \emph{CVPR Workshop}. IEEE, 2017.

\bibitem[Chu et~al.(2017)Chu, Yang, Ouyang, Ma, Yuille, and Wang]{chu2017multi}
X.~Chu, W.~Yang, W.~Ouyang, C.~Ma, A.~L. Yuille, and X.~Wang.
\newblock Multi-context attention for human pose estimation.
\newblock In \emph{Proc. {CVPR}}. IEEE, 2017.

\bibitem[Dabral et~al.(2017)Dabral, Mundhada, Kusupati, Afaque, and
  Jain]{dabral2017structure}
R.~Dabral, A.~Mundhada, U.~Kusupati, S.~Afaque, and A.~Jain.
\newblock Structure-aware and temporally coherent {3D} human pose estimation.
\newblock \emph{\arxiv{1711.09250}}, 2017.

\bibitem[He et~al.(2016)He, Zhang, Ren, and Sun]{he2016deep}
K.~He, X.~Zhang, S.~Ren, and J.~Sun.
\newblock Deep residual learning for image recognition.
\newblock In \emph{Proc. {CVPR}}. IEEE, 2016.

\bibitem[Ionescu et~al.(2014)Ionescu, Papava, Olaru, and
  Sminchisescu]{ionescu2014human3}
C.~Ionescu, D.~Papava, V.~Olaru, and C.~Sminchisescu.
\newblock {Human3.6M}: Large scale datasets and predictive methods for {3D}
  human sensing in natural environments.
\newblock \emph{{TPAMI}}, 2014.

\bibitem[Kanazawa et~al.(2018)Kanazawa, Black, Jacobs, and
  Malik]{kanazawa2017end}
A.~Kanazawa, M.~J. Black, D.~W. Jacobs, and J.~Malik.
\newblock End-to-end recovery of human shape and pose.
\newblock In \emph{Proc. {CVPR}}. IEEE, 2018.

\bibitem[Lin(1991)]{lin1991divergence}
J.~Lin.
\newblock Divergence measures based on the {Shannon} entropy.
\newblock \emph{Trans. Inf. Theory}, 1991.

\bibitem[Lin et~al.(2017)Lin, Lin, Liang, Wang, and Cheng]{lin2017recurrent}
M.~Lin, L.~Lin, X.~Liang, K.~Wang, and H.~Cheng.
\newblock Recurrent {3D} pose sequence machines.
\newblock In \emph{Proc. {CVPR}}. IEEE, 2017.

\bibitem[Luvizon et~al.(2017)Luvizon, Tabia, and Picard]{luvizon2017human}
D.~C. Luvizon, H.~Tabia, and D.~Picard.
\newblock Human pose regression by combining indirect part detection and
  contextual information.
\newblock \emph{\arxiv{1710.02322}}, 2017.

\bibitem[Luvizon et~al.(2018)Luvizon, Picard, and Tabia]{luvizon20182d}
D.~C. Luvizon, D.~Picard, and H.~Tabia.
\newblock {2D}/{3D} pose estimation and action recognition using multitask deep
  learning.
\newblock In \emph{Proc. {CVPR}}. IEEE, 2018.

\bibitem[Marinoiu et~al.(2016)Marinoiu, Papava, and
  Sminchisescu]{marinoiu2016pictorial}
E.~Marinoiu, D.~Papava, and C.~Sminchisescu.
\newblock Pictorial human spaces: A computational study on the human perception
  of {3D} articulated poses.
\newblock \emph{{IJCV}}, 2016.

\bibitem[Martinez et~al.(2017)Martinez, Hossain, Romero, and
  Little]{martinez2017simple}
J.~Martinez, R.~Hossain, J.~Romero, and J.~J. Little.
\newblock A simple yet effective baseline for {3D} human pose estimation.
\newblock In \emph{Proc. {ICCV}}. IEEE, 2017.

\bibitem[Mehta et~al.(2017{\natexlab{a}})Mehta, Rhodin, Casas, Fua,
  Sotnychenko, Xu, and Theobalt]{mehta2017monocular}
D.~Mehta, H.~Rhodin, D.~Casas, P.~Fua, O.~Sotnychenko, W.~Xu, and C.~Theobalt.
\newblock Monocular {3D} human pose estimation in the wild using improved {CNN}
  supervision.
\newblock In \emph{Proc. {3DV}}, 2017{\natexlab{a}}.

\bibitem[Mehta et~al.(2017{\natexlab{b}})Mehta, Sridhar, Sotnychenko, Rhodin,
  Shafiei, Seidel, Xu, Casas, and Theobalt]{mehta2017vnect}
D.~Mehta, S.~Sridhar, O.~Sotnychenko, H.~Rhodin, M.~Shafiei, H.-P. Seidel,
  W.~Xu, D.~Casas, and C.~Theobalt.
\newblock {VNect}: Real-time {3D} human pose estimation with a single {RGB}
  camera.
\newblock \emph{{ACM} {TOG}}, 2017{\natexlab{b}}.

\bibitem[Newell et~al.(2016)Newell, Yang, and Deng]{newell2016stacked}
A.~Newell, K.~Yang, and J.~Deng.
\newblock Stacked hourglass networks for human pose estimation.
\newblock In \emph{Proc. ECCV}. Springer, 2016.

\bibitem[Pavlakos et~al.(2017)Pavlakos, Zhou, Derpanis, and
  Daniilidis]{pavlakos2017coarse}
G.~Pavlakos, X.~Zhou, K.~G. Derpanis, and K.~Daniilidis.
\newblock Coarse-to-fine volumetric prediction for single-image {3D} human
  pose.
\newblock In \emph{Proc. {CVPR}}. IEEE, 2017.

\bibitem[Pavlakos et~al.(2018 (in press))Pavlakos, Zhou, and
  Daniilidis]{pavlakos2018ordinal}
G.~Pavlakos, X.~Zhou, and K.~Daniilidis.
\newblock Ordinal depth supervision for {3D} human pose estimation.
\newblock In \emph{Proc. {CVPR}}. IEEE, 2018 (in press).

\bibitem[Pullen and Bregler(2002)]{pullen2002motion}
K.~Pullen and C.~Bregler.
\newblock Motion capture assisted animation: Texturing and synthesis.
\newblock \emph{{ACM} {TOG}}, 2002.

\bibitem[Rhodin et~al.(2014)Rhodin, Tompkin, In~Kim, Varanasi, Seidel, and
  Theobalt]{rhodin2014interactive}
H.~Rhodin, J.~Tompkin, K.~In~Kim, K.~Varanasi, H.-P. Seidel, and C.~Theobalt.
\newblock Interactive motion mapping for real-time character control.
\newblock In \emph{CGF}. Wiley, 2014.

\bibitem[Rogez et~al.(2017)Rogez, Weinzaepfel, and Schmid]{rogez2017lcr}
G.~Rogez, P.~Weinzaepfel, and C.~Schmid.
\newblock {LCR-Net}: Localization-classification-regression for human pose.
\newblock In \emph{Proc. {CVPR}}. IEEE, 2017.

\bibitem[Shotton et~al.(2011)Shotton, Fitzgibbon, Cook, Sharp, Finocchio,
  Moore, Kipman, and Blake]{shotton2011kinect}
J.~Shotton, A.~Fitzgibbon, M.~Cook, T.~Sharp, M.~Finocchio, R.~Moore,
  A.~Kipman, and A.~Blake.
\newblock Real-time human pose recognition in parts from single depth images.
\newblock In \emph{Proc. {CVPR}}. IEEE, 2011.

\bibitem[Smith(2018)]{smith2018disciplined}
L.~N. Smith.
\newblock A disciplined approach to neural network hyper-parameters: Part
  1--learning rate, batch size, momentum, and weight decay.
\newblock \emph{\arxiv{1803.09820}}, 2018.

\bibitem[Sun et~al.(2017)Sun, Shang, Liang, and Wei]{sun2017compositional}
X.~Sun, J.~Shang, S.~Liang, and Y.~Wei.
\newblock Compositional human pose regression.
\newblock In \emph{Proc. {ICCV}}. IEEE, 2017.

\bibitem[Szegedy et~al.(2017)Szegedy, Ioffe, Vanhoucke, and
  Alemi]{szegedy2017inception}
C.~Szegedy, S.~Ioffe, V.~Vanhoucke, and A.~A. Alemi.
\newblock {Inception-v4}, {Inception-ResNet} and the impact of residual
  connections on learning.
\newblock In \emph{Proc. {AAAI}}, 2017.

\bibitem[Tekin et~al.(2017)Tekin, Marquez~Neila, Salzmann, and
  Fua]{tekin2017learning}
B.~Tekin, P.~Marquez~Neila, M.~Salzmann, and P.~Fua.
\newblock Learning to fuse {2D} and {3D} image cues for monocular body pose
  estimation.
\newblock In \emph{Proc. {ICCV}}. IEEE, 2017.

\bibitem[Tompson et~al.(2014)Tompson, Jain, LeCun, and
  Bregler]{tompson2014joint}
J.~J. Tompson, A.~Jain, Y.~LeCun, and C.~Bregler.
\newblock Joint training of a convolutional network and a graphical model for
  human pose estimation.
\newblock In \emph{Proc. {NIPS}}, 2014.

\bibitem[Toshev and Szegedy(2014)]{toshev2014deeppose}
A.~Toshev and C.~Szegedy.
\newblock {DeepPose}: Human pose estimation via deep neural networks.
\newblock In \emph{Proc. {CVPR}}. IEEE, 2014.

\bibitem[Yang et~al.(2017)Yang, Li, Ouyang, Li, and Wang]{yang2017learning}
W.~Yang, S.~Li, W.~Ouyang, H.~Li, and X.~Wang.
\newblock Learning feature pyramids for human pose estimation.
\newblock In \emph{Proc. {ICCV}}. IEEE, 2017.

\bibitem[Yi et~al.(2016)Yi, Trulls, Lepetit, and Fua]{yi2016lift}
K.~M. Yi, E.~Trulls, V.~Lepetit, and P.~Fua.
\newblock Lift: Learned invariant feature transform.
\newblock In \emph{Proc. ECCV}. Springer, 2016.

\end{thebibliography}

\end{document}